# Evaluating Large Language Models on Urdu Idioms

**Muhammad Farmal Khan[1], Mousumi Akter[1,2]**
[1]Technical University Dortmund [2]Research Center Trustworthy Data Science and Security
{farmal.khan, mousumi.akter}@tu-dortmund.de

## Abstract

Idioms remain a persistent challenge in natural language processing due to their figurative and culturally grounded meanings, which distinguish them from literal expressions. Although recent advances in large language models (LLMs) have improved idiom handling across several languages, limited attention has been given to low-resource languages such as Urdu. In this work, we present a comprehensive benchmark for Urdu–English idiomatic translation, consisting of a manually verified dataset of 4,000 aligned idiom–sentence pairs in both Perso-Arabic (native Urdu script) and Romanized Urdu. We evaluate multiple tasks, including translation, paraphrasing, idiom span detection, and back-translation, using diverse prompting strategies such as cultural prompting, idiomatic prompting, and few-shot learning. Our findings show that frontier LLMs consistently outperform traditional neural machine translation systems across all evaluation settings, particularly in preserving figurative and metaphorical meaning. While models demonstrate relatively stable performance on native Urdu script, the absence of standardized orthography in Romanized Urdu introduces substantial challenges for consistency and idiom span detection. This work establishes a high-quality benchmark for cross-script idiomatic evaluation in Urdu and underscores the importance of prompt engineering in preserving figurative language meaning across languages.[1]

## 1 Introduction

Idiomatic expressions pose a long-standing challenge in machine translation due to their non-compositional semantics, cultural grounding, and contextual dependence. For example, the Urdu idiom اونٹ کے منہ میں زیرہ ("a cumin seed in a camel's mouth") corresponds to the English idiom "a drop in the ocean," illustrating how figurative meaning is easily lost when systems rely on literal alignments. Accurate translation requires recognizing idiomatic spans, interpreting figurative intent, and generating culturally appropriate equivalents rather than surface-level mappings.

Urdu, one of the most widely spoken languages globally (Eberhard et al., 2023), presents additional complexity due to its dual writing systems: the formal Perso–Arabic script (Native Urdu) and the widely used but orthographically inconsistent Roman Urdu. The lack of standardized spelling and phonetic ambiguity in Roman Urdu complicate idiom preservation across scripts. Despite the prevalence of idioms in Urdu communication, there is limited research on how modern LLMs or NMT systems handle idiomatic content in either script, and no unified evaluation spanning translation, paraphrasing, span detection, and back-translation.

To address these gaps, we introduce the first comprehensive benchmark for Urdu–English idiom understanding and translation. Our benchmark extends prior idiom-focused work by covering multiple subtasks, including direct and literal translation, culturally grounded translation, paraphrasing, idiom-span detection, idiomatic back-translation, and few-shot translation with 5-shot and 10-shot demonstrations. This design enables analysis of idiom recognition, cross-script transfer, prompt sensitivity, and robustness under varied conditions.

We evaluate seven open- and closed-source LLMs alongside two NMT baselines on 4,000 idiom–sentence pairs split equally across scripts, using a diverse metric suite spanning lexical (BLEU, ChrF), semantic (BERTScore, COMET, XCOMET), and embedding-based (LASER, LaBSE, USE-multilingual) similarity measures.

**Our main contributions are:**

- We release a unified benchmark for Urdu–

[1]Code and dataset are publicly available at: https://github.com/Mousumi44/Urdu-Idioms

English idiom understanding across translation, paraphrasing, idiom-span detection, and back-translation in Native and Roman Urdu.

- We provide a systematic analysis of prompt sensitivity in idiom translation for Urdu using eight prompting strategies.
- We present a cross-script comparison of idiomatic translation quality, revealing substantial performance gaps driven by orthographic representation.
- We introduce a comprehensive human and multi-metric evaluation combining BLEU, ChrF, BERTScore, COMET, XCOMET, ROUGE, Levenshtein distance, and embedding similarity.

## 2 Related Work

**Idiom Comprehension in NLP.** A growing body of research has developed annotated resources for idiom identification, comprehension, and translation across languages. Early work introduced idiom resources for Hindi and Marathi (Singh et al., 2016), followed by dedicated datasets for English (Zhou et al., 2021), the multilingual Id10M benchmark (Tedeschi et al., 2022), and resources targeting Chinese idioms (Qiang et al., 2023). Additional multilingual resources have expanded idiom research to several Indian languages (Agrawal et al., 2018) and to languages including Catalan, Italian, and Russian (Sentsova et al., 2025). More recently, the PARSEME 2.0 Shared Task (Scholivet et al., 2026) has further advanced multilingual research on idiom identification and paraphrasing across 14 languages.

Idiomatic translation has been examined in several language pairs, with prior studies highlighting the difficulty of preserving figurative meaning in neural systems. Fadaee et al. (Fadaee et al., 2018) showed that data augmentation and semantic cues can improve idiom handling in NMT, while Liu et al. (Liu et al., 2023) explored cross-lingual idiom understanding using multilingual encoders. Recent work on Persian-English idiom translation demonstrates that retrieval-augmented prompting and semantic similarity models can enhance figurative meaning retention (Rezaeimanesh et al., 2025). In contrast, research on Urdu NLP has predominantly addressed broader low-resource challenges, including morphological complexity, script variation, and orthographic inconsistency (Furqan et al., 2024; Butt et al., 2025).

**Research Gap and Our Contribution** Despite the prevalence of idioms in Urdu communication, no prior work has investigated idiomatic translation across Native and Roman Urdu, and existing machine translation benchmarks do not include idiom-specific evaluation for Urdu. This work addresses this gap by introducing the first benchmark dedicated to Urdu–English idiomatic translation and evaluation.

## 3 Dataset

We construct a unified dataset of 4,000 Urdu–English idiom–sentence pairs, divided equally across writing systems. Each instance contains an Urdu sentence with at least one idiomatic expression, the extracted idiom, and a gold English translation verified by bilingual annotators. The dataset covers a broad range of figurative expressions spanning common proverbs, culture-dependent idioms, and semantically opaque phrases.

Table 1 shows an example idiom and its gold translation.

### 3.1 Native Urdu

The Native Urdu dataset contains 2,000 sentences written in the Perso–Arabic script, commonly used in formal communication, print media, and education. Idioms were collected from publicly available resources such as UrduPod101,[2] LingApp,[3] and EnglishUms[4], supplemented with manually curated examples from native speakers. Each idiom was paired with an English equivalent and contextual sentence. The initial English equivalents and contextual sentences were drafted with assistance from GPT-5; however, these outputs served only as preliminary templates. A panel of three bilingual postgraduate researchers (including the author) manually reviewed, edited, and corrected every entry to ensure absolute semantic fidelity and idiomatic resonance. Concretely, annotators post-edited GPT-5 outputs rather than simply discarding low-quality entries: each sentence was iteratively revised until all three reviewers reached consensus on its accuracy, naturalness, and idiomatic appropriateness. Entries for which consensus could not be reached were discarded and replaced with new examples.

[2] https://www.urdupod101.com/
[3] https://ling-app.com/
[4] https://englishums.com/

**Idiom (Native Urdu):** اونٹ کے منہ میں زیرہ

**Idiom (Roman Urdu):** Oont ke munh mein zeera

**Idiom (English):** A drop in the ocean

**Sentence (Native Urdu):** یہ تنخواہ تو میرے اخراجات کے مقابلے میں اونٹ کے منہ میں زیرہ ہے۔

**Sentence (Roman Urdu):** Ye tankhwah to mere kharchon ke muqablay mein oont ke munh mein zeera hai.

**Gold Translation:** This salary is just a drop in the ocean compared to my expenses.

Table 1: Example instance from the Native and Roman Urdu datasets.

### 3.2 Roman Urdu

Roman Urdu refers to the representation of Urdu using the Latin script, commonly used in informal digital communication such as texting, social media, and online forums. Unlike Native Urdu, Roman Urdu lacks standardized orthography, resulting in substantial spelling variation and increased lexical diversity, which pose challenges for translation and evaluation systems.

For this work, the Roman Urdu portion of the dataset was constructed to mirror the Native Urdu split. Starting from the 2,000 Native Urdu idioms and their accompanying sentences, we utilized GPT-5[5] to produce the initial transliterated drafts. To ensure the quality of this synthetic starting point, every entry was subjected to a thorough manual review by three native speakers.

These reviewers *post-edited* the GPT-5 transliterations where each entry was corrected and refined until it reflected natural, real-world usage patterns, including the informal phonetic spellings typical of digital Urdu. Entries for which consensus on naturalness and accuracy could not be reached were discarded. The final Roman Urdu dataset consists of 2,000 high-quality idiom–sentence pairs aligned with their Native Urdu and English counterparts, enabling controlled cross-script comparisons in downstream evaluation.

### 3.3 Dataset Statistics

Table 2 presents a quantitative comparison of the curated Native Urdu and Roman Urdu idiom–sentence datasets. Each split contains 2,000 idiom–sentence pairs, resulting in a total of 4,000 aligned instances with verified English translations.

Both datasets exhibit similar structural patterns: idioms average between six and seven words, while sentences typically range from nine to ten words. Roman Urdu shows a slightly larger vocabulary size (3,254 vs. 3,033 in Native Urdu) as well as a higher Type–Token Ratio (TTR), reflecting its greater orthographic variability and lack of standardized spelling conventions.

Figures 1 and 2 visualize the distributions of idiom and sentence lengths, respectively.

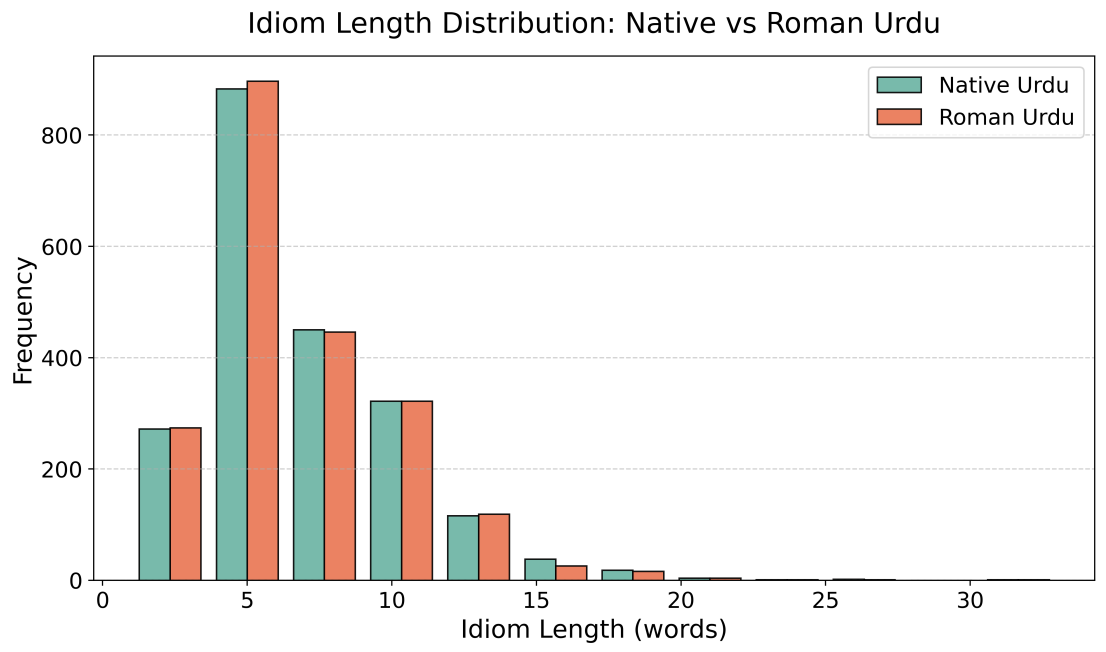

Figure 1: Distribution of idiom lengths in the Native and Roman Urdu datasets.

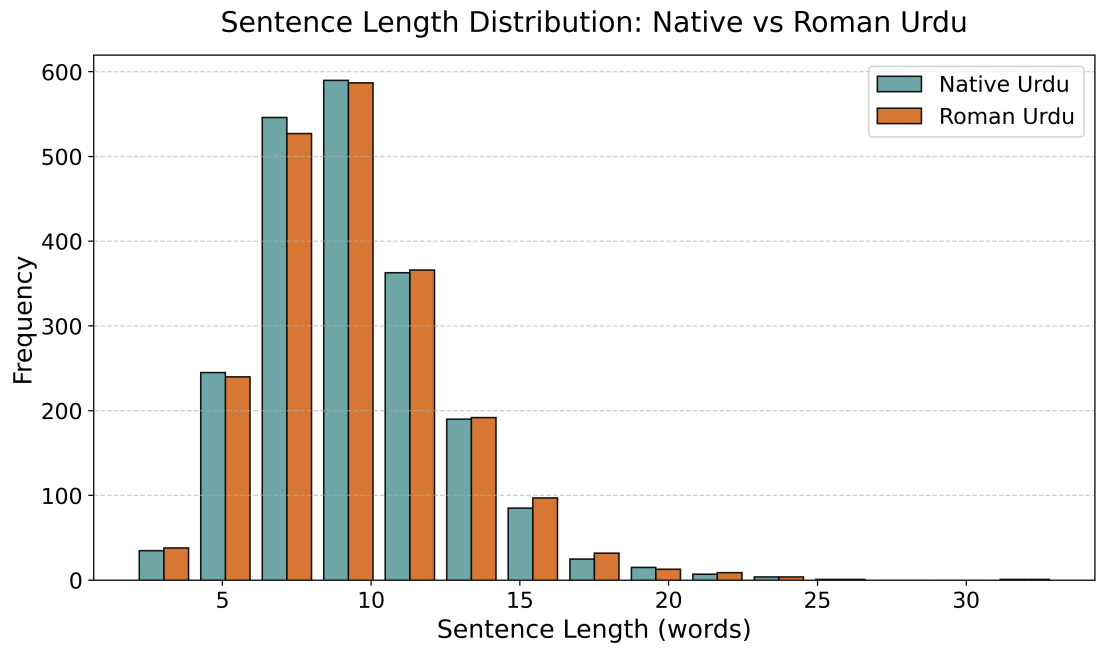

Figure 2: Distribution of sentence lengths in the Native and Roman Urdu datasets.

## 4 Experimental Design

This section outlines the prompting strategies, model configurations, and evaluation metrics used across all experiments, including idiom translation, idiom classification, English–Urdu back-translation, and Native-Roman Urdu comparison.

### 4.1 Prompting Strategies

Table 3 shows the set of prompting strategies used in the experiments.

All large language model experiments were carried out using a chat-based interface in which the

[5] https://chat.openai.com/

| Dataset | # of Pairs | Avg. Idiom Length | Avg. Sentence Length | Vocab. Size | TTR |
|---|---|---|---|---|---|
| Native Urdu | 2000 | 6.74 | 9.62 | 3033 | 0.213 |
| Roman Urdu | 2000 | 6.65 | 9.71 | 3254 | 0.232 |

Table 2: Linguistic statistics of the curated Urdu idiom translation datasets.

| Prompt | Prompt Instruction |
|---|---|
| Literal Translation | Translate the following Urdu text to English. Do not include explanations, reasoning, or additional text. |
| Cultural Translation | Translate the following Urdu text into English. Use natural expressions and preserve the cultural meaning. Do not include explanations, reasoning, or additional text. |
| Paraphrase Translation | Paraphrase the following Urdu sentence into fluent English, keeping the meaning intact. Avoid word-for-word translations. Do not include explanations, reasoning, or additional text. |
| Idiomatic Translation | Identify the idiom in this Urdu sentence and produce an equivalent English idiom. Do not translate word-for-word. Do not include explanations or reasoning. |
| Idiomatic Back Translation | Translate the following Urdu idiom to an equivalent English idiom, and then back to Urdu. Provide only the final back-translated Urdu idiom. |
| Idiom Span Detection | Detect and extract the Urdu idiom in the sentence. Provide only the final detected Urdu idiom. |
| Few-Shot Translation | Translate the following Urdu text to English using the examples as references. |

Table 3: Summary of full prompts used in the experiments.

system role was explicitly set to *"You are an expert translator fluent in both Urdu and English."*

The Few-Shot translation prompt contains two variations: 5-Shot and 10-Shot with 5 and 10 Urdu-English translated examples provided in the prompts respectively.

**Few-Shot Example Selection.** To prevent in-context contamination, the few-shot examples were drawn exclusively from a held-out pool that is entirely disjoint from the 4,000-instance test set used in all evaluations. For each iteration, examples were selected randomly without replacement from this held-out pool. No similarity-based retrieval was applied; examples were not chosen based on semantic proximity to the test instance. The same sampling procedure was applied consistently across all models to ensure fair comparison. This design ensures that performance gains observed in few-shot settings reflect genuine in-context learning from representative idiom–translation pairs rather than any overlap with test inputs.

### 4.2 Models

We evaluate a mix of closed-source and open-source LLMs and Neural Machine Translation (NMT) models. The LLMs include Gemini-3 [6], GPT-OSS-20B (Kumar et al., 2025), Qwen 3-8B (Yang et al., 2025), DeepSeek R1-Distill-Llama-8B (Guo et al., 2025), Gemma 3-1B (Team et al., 2025), Mistral Instruct-7B (Jiang et al., 2023), and LLama-3.2-3B-Instruct (Grattafiori et al., 2024). For NMT baselines, nllb-200-3.3B, nllb-200-distilled-600M (Team et al., 2022), and Google Translate [7].

The model temperature was set to 0.8 with a maximum output limit of 500 tokens. To account for models that generated conversational fillers, reasoning, or explanations alongside the translation, we employed a multi-step regular expression (regex) pipeline to programmatically isolate and extract the final translated output for evaluation. The extraction procedure applied the following steps in order: (1) remove chain-of-thought reasoning blocks enclosed in `<think>...</think>` tags, as produced by reasoning-oriented models such as DeepSeek-R1; (2) strip surrounding quotation marks ("", '', «»), leading and trailing whitespace, and markdown formatting markers (e.g., `**bold**`).

[6] https://gemini.google.com/
[7] https://translate.google.com/

### 4.3 Human Evaluation

To establish a gold-standard benchmark, we conducted a human evaluation of 1,000 translations (100 samples per model) generated from the *Cultural Translation* experiment. The dataset consisted of 500 Native Urdu and 500 Roman Urdu sentences. Three MSc students with native Urdu proficiency and strong command of English volunteered for the evaluation; who hold postgraduate qualifications and have extensive experience in Urdu–English bilingual contexts, making them well-suited for assessing figurative and cultural translation quality. The annotators were blinded to system identity using anonymous labels (System A–J) with randomised sample order per annotator.

Each translation was evaluated on a 5-point Likert scale across three core dimensions:

- **Semantic Accuracy:** Evaluates whether the translation captures the underlying figurative and metaphorical meaning of the idiom rather than providing a literal rendering.
- **Fluency:** Assesses the grammatical correctness, naturalness, and readability of the output within an English linguistic context.
- **Cultural Equivalence:** Measures how well the translation resonates as a functional equivalent in English, determining if the output feels like a natural expression or idiom rather than an explained concept.

The final *Human Score* for each sample was calculated as the arithmetic mean of these dimensions.

### 4.4 Automated Evaluation Metrics

We employ a combination of lexical, semantic, and idiomatic evaluation metrics to comprehensively assess translation quality. BLEU (Papineni et al., 2002) and ChrF (Popović, 2015) measure surface-level n-gram overlap between the model output and reference translations. BERTScore (Zhang et al., 2020), COMET (Rei et al., 2020), and XCOMET (Guerreiro et al., 2024) capture semantic similarity and cross-lingual adequacy. ROUGE (Lin, 2004) is used for paraphrase evaluation, while Levenshtein distance and unigram overlap assess idiom-level variation and lexical preservation. Additionally, cosine similarity computed using multilingual embeddings (LASER, LaBSE, and USE-multilingual) quantifies semantic fidelity across scripts.

## 5 Results

### 5.1 Human Evaluation Results

To assess the reliability of the human evaluation, we computed inter-annotator agreement using Fleiss' kappa ($\kappa$). Agreement was calculated separately for each evaluation dimension, namely *Semantic Accuracy*, *Fluency*, and *Cultural Equivalence*, as well as for the aggregated *Human Score*. Table 4 reports the resulting $\kappa$ values.

| Evaluation Dimension | Fleiss' $\kappa$ |
|---|---|
| Semantic Accuracy | 0.66 |
| Fluency | 0.75 |
| Cultural Equivalence | 0.71 |
| Human Score | **0.68** |

Table 4: Inter-annotator agreement measured using Fleiss' kappa across evaluation dimensions.

The inter-annotator agreement results in Table 4 demonstrate substantial consensus across all dimensions. The overall Human Score achieved a Fleiss' $\kappa$ of 0.68, while Fluency reached the highest alignment at 0.75. Semantic Accuracy (0.66) and Cultural Equivalence (0.71) also showed consistent substantial agreement. These robust values validate the human evaluation as a reliable ground truth, confirming that the scoring rubric was applied consistently across evaluators.

A representative example of the evaluation scoring process is provided in Table 5, demonstrating how a model's literal translation is penalized for semantic and cultural divergence despite maintaining grammatical fluency. The final *Human Score* for each sample was calculated as the arithmetic mean of these three dimensions.

Table 6 reports pairwise unweighted Cohen's $\kappa$ between each pair of annotators across all evaluation dimensions. Annotators are anonymised as A, B, and C. All pairs achieve substantial agreement across every dimension, confirming that the rating rubric was applied consistently regardless of which annotator pair is considered.

### 5.2 Correlation Analysis

To assess the reliability of automated metrics for Urdu idiomatic translation, we calculated Spearman's rank correlation coefficient ($\rho$) between the automated scores and the averaged human ratings across 10 models.

Table 5: Example of the Human Evaluation Scoring Process for a Model Output.

| Field | Content |
|---|---|
| **Source (Native Urdu)** | اونٹ کے منہ میں زیرہ |
| **Source (Roman)** | Oont ke munh mein zeera |
| **Reference (Gold)** | A drop in the ocean |
| **Model Output** | *"Cumin in a camel's mouth"* |
| **Dimension** | **Annotator Score (1–5) & Justification** |
| Semantic Accuracy | **1/5** – The literal translation fails to convey the intended meaning of inadequacy or scarcity. |
| Fluency | **4/5** – The sentence is grammatically correct English, though semantically nonsensical in context. |
| Cultural Equivalence | **1/5** – There is no attempt to find a corresponding English metaphor; the cultural anchor is lost. |
| **Final Aggregated Score** | **2.0 / 5.0** |

| **Dimension** | **A–B** | **A–C** | **B–C** | **Avg.** |
|---|---|---|---|---|
| Semantic Accuracy | 0.74 | 0.63 | 0.61 | 0.66 |
| Fluency | 0.81 | 0.72 | 0.70 | 0.74 |
| Cultural Equivalence | 0.73 | 0.69 | 0.67 | 0.70 |
| Human Score | 0.76 | 0.64 | 0.62 | 0.67 |

Table 6: Pairwise Cohen's $\kappa$ between annotator pairs across evaluation dimensions. Avg. denotes the mean across the three pairs.

As shown in Figure 3, neural-based metrics such as COMET and XCOMET consistently outperform traditional surface-matching metrics like BLEU and chrF. For instance, Gemini-3 achieves a $\rho$ of $0.845$ with COMET but only $0.166$ with BLEU ($p > 0.05$), highlighting BLEU's inability to capture the figurative nuances of idioms.

*The correlation matrix of the metrics themselves reveals a strong alignment between COMET and XCOMET*, whereas BLEU shows a weak or negligible relationship with human judgment in several high-performing models (e.g., DeepSeek, $\rho = -0.04$). These results suggest that semantic-aware metrics are essential for evaluating culturally dense tasks like idiomatic translation.

### 5.3 Translation Performance and Strategy Analysis

Tables 7 and 8 reveal a consistent hierarchy where frontier LLMs outperform NMT models across both scripts. As established in Section 3, COMET and XCOMET are the primary quality indicators ($\rho = 0.845$), while BLEU remains negligible (<0.05) and uncorrelated with idiomatic fidelity.

In Native Urdu, Gemini-3 leads with XCOMET 0.77 and COMET 0.68, followed by GPT-OSS-20B (XCOMET 0.74). Performance drops in Roman Urdu due to orthographic inconsistency, though Gemini-3 retains the strongest relative performance (XCOMET 0.70). NMT models suffer the steepest declines: NLLB's ChrF falls from 21.14 to 9.65, underscoring LLMs' greater robustness to transliterated input.

The strategy analysis in Table 8 confirms that prompting methodology is as critical as model architecture. The Direct NMT approach struggled across both scripts, particularly in Roman Urdu where it recorded its lowest semantic score (XCOMET: 0.49). This reinforces the finding that standard pipelines are not equipped for the combined complexity of non-standardized scripts and idiomatic metaphors. Transitioning from Literal to Cultural prompting resulted in a consistent performance lift. In Native Urdu, Cultural Prompting achieved an XCOMET of 0.72, outperforming the Literal approach (0.66). By explicitly instructing models to seek cultural equivalents rather than literal translations, the models activate broader semantic associations that better preserve the idiom's intent. Even in the challenging Roman Urdu environment, the Cultural strategy maintained the lead (XCOMET: 0.57), proving that cultural instruction is a viable mechanism for mitigating semantic loss in transliterated Urdu.

### 5.4 Comparison of Few-Shot and Zero-Shot Translations

To evaluate the influence of in-context learning on idiomatic Urdu–English translation, we compare two few-shot settings (5-shot and 10-shot) with two representative zero-shot prompting strategies (Literal and Cultural). Table 9 reports BERTScore, COMET, XCOMET, BLEU, and ChrF scores for all settings.

Few-shot prompting consistently improves idiomatic translation quality over zero-shot settings, with semantic-aware metrics showing the most substantial gains. COMET and XCOMET, which correlate most strongly with human judgment, rise from zero-shot peaks of 0.64 and 0.67 to few-shot highs of 0.72 and 0.81, respectively. This progression indicates that in-context examples significantly enhance cross-lingual meaning preservation and idiomatic fidelity. Interestingly, while the move from 5-shot to 10-shot yields a notable increase in XCOMET, lexical metrics such as BLEU and ChrF remain stagnant or slightly decline. These results suggest that increasing the shot size primarily refines deep semantic nuance rather than surface-level lexical alignment, with diminishing returns for string-matching accuracy beyond five examples.

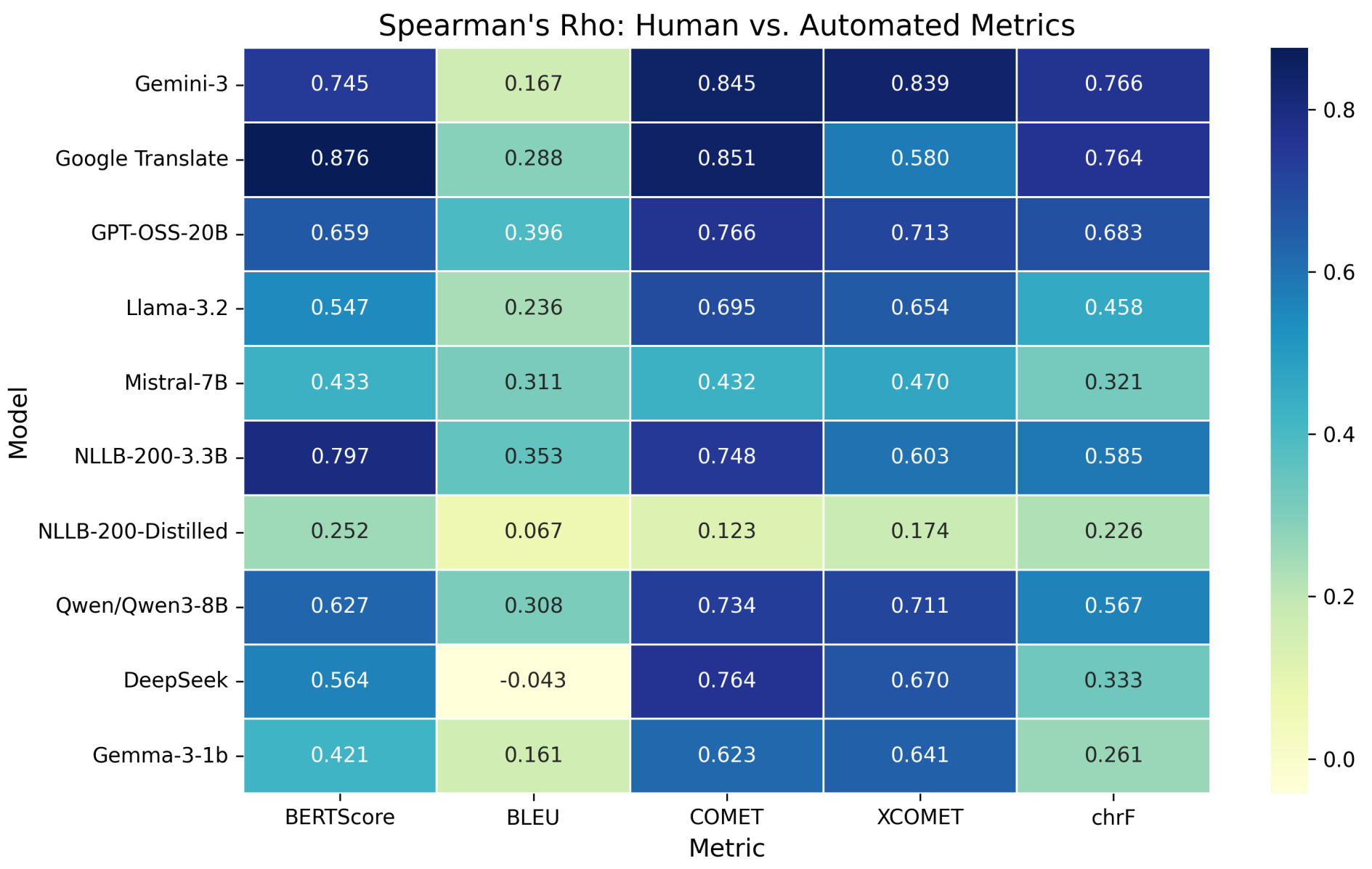

Figure 3: Heatmap of Spearman's Rank Correlation ($\rho$) between human evaluations and automated metrics across ten models. Darker colors indicate stronger alignment with human judgment.

| | Native Urdu | | | | | Roman Urdu | | | | |
|---|---|---|---|---|---|---|---|---|---|---|
| **Model** | **BERT** | **CMT** | **XCMT** | **BLEU** | **ChrF** | **BERT** | **CMT** | **XCMT** | **BLEU** | **ChrF** |
| Qwen 3-8B | 0.88 | 0.65 | 0.71 | 0.03 | 25.38 | 0.83 | 0.60 | 0.62 | 0.04 | 23.79 |
| DeepSeek-8B | 0.87 | 0.59 | 0.59 | 0.02 | 21.82 | 0.81 | 0.52 | 0.46 | 0.02 | 20.08 |
| NLLB 3.3B | 0.87 | 0.63 | 0.61 | 0.02 | 21.14 | 0.82 | 0.41 | 0.34 | 0.00 | 9.65 |
| NLLB 600M | 0.87 | 0.64 | 0.62 | 0.02 | 21.33 | 0.78 | 0.42 | 0.56 | 0.00 | 10.25 |
| Gemini-3 | **0.91** | **0.68** | **0.77** | 0.16 | 42.08 | **0.86** | **0.67** | **0.70** | 0.12 | 35.95 |
| Gemma 3-1B | 0.87 | 0.59 | 0.57 | 0.01 | 20.02 | 0.79 | 0.46 | 0.36 | 0.01 | 13.81 |
| Google Trans. | 0.86 | 0.63 | 0.64 | 0.03 | 21.70 | 0.80 | 0.44 | 0.56 | 0.00 | 12.26 |
| LLaMA 3.2-3B | 0.88 | 0.60 | 0.64 | 0.02 | 21.64 | 0.83 | 0.56 | 0.50 | 0.04 | 23.83 |
| Mistral 7B | 0.87 | 0.57 | 0.56 | 0.01 | 19.43 | 0.81 | 0.55 | 0.56 | 0.02 | 21.78 |
| GPT-OSS-20B | 0.89 | 0.66 | 0.74 | 0.04 | 26.33 | 0.84 | 0.62 | 0.66 | 0.06 | 28.37 |

Table 7: Comparative performance across Native and Roman Urdu scripts. The best scores for semantic metrics (COMET and XCOMET) are highlighted in bold.

### 5.5 Paraphrasing Experiment

To evaluate how well models generate fluent English paraphrases of Urdu idioms and sentences while preserving meaning, we assess seven LLMs using semantic similarity, surface-level lexical overlap, Levenshtein distance, and multilingual embedding coherence. Table 10 presents the results.

Across evaluation metrics, Gemini-3 and GPT-OSS-20B consistently outperform smaller models. Gemini-3 achieves the highest lexical alignment with a ROUGE score of 0.32 and a unigram overlap of 35.94%, while maintaining top-tier semantic fidelity (BERTScore 0.90). Levenshtein distances between 39 and 43 across all systems suggest substantial rewriting behavior, with Gemini-3 and GPT-OSS-20B demonstrating the most precise paraphrasing capabilities. Cross-lingual embedding-based evaluations using LASER, LaBSE, and multilingual USE further confirm that Gemini-3 leads in preserving deep semantic structure (LaBSE 0.59), whereas smaller models show weaker alignment.

### 5.6 Idiom Span Detection Results

In addition to translation and paraphrasing, we evaluated the models on their ability to detect idiomatic expressions within Urdu sentences. This experiment measures how accurately models can identify the span of an idiom, which is critical for subsequent translation or paraphrasing tasks.

| Experiment | Native Urdu BERT | CMT | XCMT | BLEU | ChrF | Roman Urdu BERT | CMT | XCMT | BLEU | ChrF |
|---|---|---|---|---|---|---|---|---|---|---|
| Direct (NMT) | 0.87 | 0.63 | 0.62 | 0.02 | 25.79 | 0.80 | 0.43 | 0.49 | 0.00 | 10.72 |
| Literal Prompt | 0.89 | 0.64 | 0.66 | 0.02 | 26.39 | 0.88 | 0.58 | 0.56 | 0.03 | 24.04 |
| Cultural Prompt | 0.89 | 0.66 | 0.72 | 0.03 | 27.65 | 0.89 | 0.60 | 0.57 | 0.04 | 25.45 |

Table 8: Comparative performance of translation strategies across Native and Roman Urdu datasets: direct NMT, literal prompting, and cultural prompting.

| Experiment | BERT | COMET | XCOMET | BLEU | ChrF |
|---|---|---|---|---|---|
| **Few-Shot Experiments** | | | | | |
| 5-Shot | 0.91 | 0.69 | 0.75 | 0.15 | 41.14 |
| 10-Shot | 0.91 | 0.72 | 0.81 | 0.14 | 41.11 |
| **Zero-Shot Experiments** | | | | | |
| Literal Translation | 0.89 | 0.64 | 0.66 | 0.02 | 24.69 |
| Cultural Translation | 0.89 | 0.64 | 0.67 | 0.03 | 25.36 |

Table 9: Comparison of few-shot and zero-shot idiomatic translation performance across semantic and lexical metrics.

| Model | BERT | ROUGE | Lev Dist | Uni Overlap | Cosine Similarity LASER | USE | LaBSE |
|---|---|---|---|---|---|---|---|
| Qwen3-8B | 0.90 | 0.30 | 39.36 | 30.42 | 0.56 | 0.53 | 0.55 |
| DeepSeek-8B | 0.89 | 0.26 | 42.34 | 27.57 | 0.49 | 0.48 | 0.51 |
| Gemma-3-1B | 0.89 | 0.21 | 42.67 | 21.21 | 0.45 | 0.43 | 0.46 |
| Llama-3.2-3B | 0.89 | 0.26 | 40.65 | 27.02 | 0.52 | 0.47 | 0.51 |
| Mistral-7B | 0.88 | 0.23 | 42.18 | 22.36 | 0.44 | 0.42 | 0.45 |
| GPT-OSS-20B | 0.90 | 0.31 | 40.36 | 32.45 | 0.57 | 0.54 | 0.57 |
| Gemini-3 | 0.90 | 0.32 | 42.41 | 35.94 | 0.58 | 0.53 | 0.59 |

Table 10: Paraphrasing results using BERTScore, ROUGE, Levenshtein distance, unigram overlap, and embedding similarities.

For evaluation, we used ROUGE to measure n-gram overlap between predicted and reference spans, Levenshtein distance and unigram overlap for exact lexical matches, and cosine similarity using multilingual embeddings to capture semantic similarity across scripts.

Gemini-3 and GPT-OSS-20B show the highest BERTScore and cosine similarity, suggesting strong semantic alignment with reference idioms. Notably, Gemini-3 achieves the lowest Levenshtein distance (12.37) and highest unigram overlap (67.00), indicating superior boundary detection. LLama-3.2B also performs well in ROUGE-based overlap. Lower scores for Mistral-7B and Gemma-3-1B indicate significant challenges in identifying idiomatic spans accurately. These results highlight a clear performance gap, where high detection accuracy directly facilitates better downstream idiomatic translation quality.

### 5.7 Qualitative Analysis

To illustrate the behavior of our prompting strategies, we provide a representative example from the dataset where the Urdu sentence اچانک خبر سنتے ھی اس کے ھاتھ پیر پھول گئے۔ is paired with the golden English translation: “She completely panicked the moment she heard the sudden news.” The idiom ھاتھ پیر پھول جانا literally means “hands and feet swelling” but idiomatically conveys sudden panic or overwhelming anxiety, making it a strong test case for figurative meaning preservation.

Table 12 demonstrates how different prompting strategies influence the model’s output, ranging from literal physical descriptions to culturally nuanced idiomatic equivalents. This comparison highlights the model’s ability to shift from surface-level translation to deep semantic interpretation when explicitly guided by task-specific instructions, and underscores why prompt design is as consequential as model selection for idiomatic translation quality.

| Model | BERT | ROUGE | Lev Dist | Uni Overlap | Cosine Similarity | | |
|---|---|---|---|---|---|---|---|
| | | | | | LASER | USE | LaBSE |
| Qwen3-8B | 0.83 | 0.48 | 15.93 | 49.57 | 0.70 | 0.62 | 0.66 |
| DeepSeek-8B | 0.73 | 0.27 | 27.18 | 27.18 | 0.54 | 0.43 | 0.48 |
| Gemma-3-1B | 0.68 | 0.15 | 38.21 | 17.06 | 0.49 | 0.43 | 0.46 |
| Llama-3.2-3B | 0.82 | 0.55 | 19.62 | 57.35 | 0.69 | 0.61 | 0.66 |
| Mistral-7B | 0.66 | 0.05 | 36.64 | 5.83 | 0.45 | 0.40 | 0.45 |
| GPT-OSS-20B | 0.84 | 0.53 | 15.67 | 53.37 | 0.72 | 0.66 | 0.69 |
| Gemini-3 | 0.88 | 0.52 | 12.37 | 67.00 | 0.71 | 0.68 | 0.65 |

Table 11: Idiom Span Detection results using ROUGE, Levenshtein distance, unigram overlap, and embedding similarities.

| Prompt | Translation Result |
|---|---|
| Literal | Her hands and feet swelled up. |
| Cultural | She suddenly became extremely nervous. |
| Paraphrase | She lost her composure and couldn't think straight. |
| Idiomatic | She completely freaked out. |
| Back Translation | وہ گھبراھٹ میں آ گئی۔ |
| Span Detection | ھاتھ پیر پھول جانا |

Table 12: Sample translation results using Gemini-3 across different prompts.

## 6 Conclusion

We present the first systematic benchmark for Urdu–English idiomatic translation across Native and Roman Urdu, covering translation, paraphrasing, idiom span detection, and back-translation. Central to this study is our curated dataset of 4,000 aligned idiom–sentence pairs, which provides a high-quality foundation for cross-script comparisons through manual verification by native speakers. Our findings indicate that large LLMs, particularly Gemini-3, GPT-OSS-20B, and Qwen 3-8B, consistently outperform NMT systems.

We found that cultural and idiomatic prompts yield the highest semantic fidelity by activating broader associations that better preserve the idiom's intent. Few-shot prompting provides additional gains, confirming that in-context learning further improves idiomatic translation quality over zero-shot settings. Cross-script analysis reveals that while semantic metrics like COMET and XCOMET remain relatively stable across scripts, the lack of standardized orthography and phonetic ambiguity in Roman Urdu continues to pose challenges for model robustness, with lexical metrics such as BLEU and ChrF showing the steepest cross-script degradation. Furthermore, our idiom span detection results emphasize that the accurate identification of figurative expressions is a critical prerequisite for high-quality downstream translation. Overall, this study offers a unified multi-metric evaluation framework for idiomatic Urdu–English translation, providing insights for LLM development, prompt engineering, and the broader challenge of figurative language processing in low-resource settings.

## Limitations

While this study provides a comprehensive benchmark for idiomatic translation, several limitations should be acknowledged. The dataset focuses exclusively on the Urdu–English language pair. Although this enables an in-depth analysis of these languages, the findings may not directly generalize to other low-resource languages or different linguistic families. The dataset comprises 4,000 idiom–sentence pairs. Although the dataset was carefully validated by multiple native speakers, its partially AI-generated template remains a potential limitation, as it may introduce evaluation bias that favors models similar to the one used during data generation. Moreover, the sample may not fully capture the diversity and variability of Urdu idioms in real-world usage. The lack of standardized orthography in Roman Urdu introduces phonetic ambiguity and spelling inconsistencies, posing challenges for both idiom detection and automated evaluation. Human evaluation was conducted exclusively under the Cultural Translation prompting condition; extending human evaluation to other strategies, such as Literal Translation and Few-Shot prompting, would enable direct comparison of model performance across different conditions and represents an important direction for future work. In addition, idiom span detection is evaluated using lexical-overlap metrics, such as ROUGE and unigram overlap, which do not adequately account for spelling variation in Roman Urdu. Future work should incorporate phonetic normalization or fuzzy string matching to enable more robust and fair cross-script span detection evaluation. Finally, the current study is limited to Urdu–English translation, and extending the benchmark to other low-resource languages remains an important direction for assessing the broader applicability and generalizability of the proposed evaluation framework.

## Acknowledgements

This work has been supported by the Research Center Trustworthy Data Science and Security `https://rc-trust.ai`, one of the Research Alliance centers within the `https://uaruhr.de`.

# A Appendix

## A.1 Resources

All experiments were performed on machines equipped with NVIDIA A100 40GB GPUs.

## A.2 Roman Urdu Pitfalls

Roman Urdu often leads to lower translation quality compared to Native Urdu due to inconsistent spelling, phonetic ambiguity, and lack of standardization. For example, the Urdu idiom ہاتھ پیر پھول جانا can appear as *haath pair phool jana*, *hath pair phool jana*, or even *haath pair phool ja na*, causing models to misinterpret or fail to detect idiomatic meaning.

Other idioms show even greater variation in Romanization:

- آسمان سے باتیں کرنا (to daydream) → *aasman se baatein karna / asman sy baatein krna*
- نرگسیت میں ڈوب جانا (to become self-absorbed) → *nargisiyat mein doob jana / nargisiat me doob jana*

These variations increase lexical ambiguity and make it harder for models to match idioms with their English equivalents.

### A.3 Comparative Analysis: Native vs. Roman Urdu

We analyze the impact of orthographic representation on model performance by comparing Native Urdu (Perso-Arabic) against Roman Urdu (transliterated Latin) across core translation tasks. Figure 4 and Figure 5 provide a holistic visualization of the ”Script Gap” using neural evaluation metrics, which prioritize semantic adequacy over surface-level string matching.

The empirical results confirm that Native Urdu yields consistently superior semantic fidelity across all neural benchmarks. As illustrated in the radar charts, the performance delta is most pronounced in tasks requiring high-precision mapping, such as *Direct Translation* and *Idiomatic Translation*. The visual ”contraction” of the Roman Urdu series relative to the Native baseline represents the performance tax imposed by phonetic noise and subword fragmentation during tokenization.

These findings reinforce the conclusion that while modern Large Language Models (LLMs) act as a robust semantic anchor for informal Urdu, Native Urdu remains the gold standard for achieving peak semantic adequacy in culturally dense, non-compositional translation tasks. The narrowest gaps observed in *Literal* and *Cultural* prompting suggest that instruction-tuning provides a degree of resilience against script-based degradation that is less present in traditional machine translation frameworks.

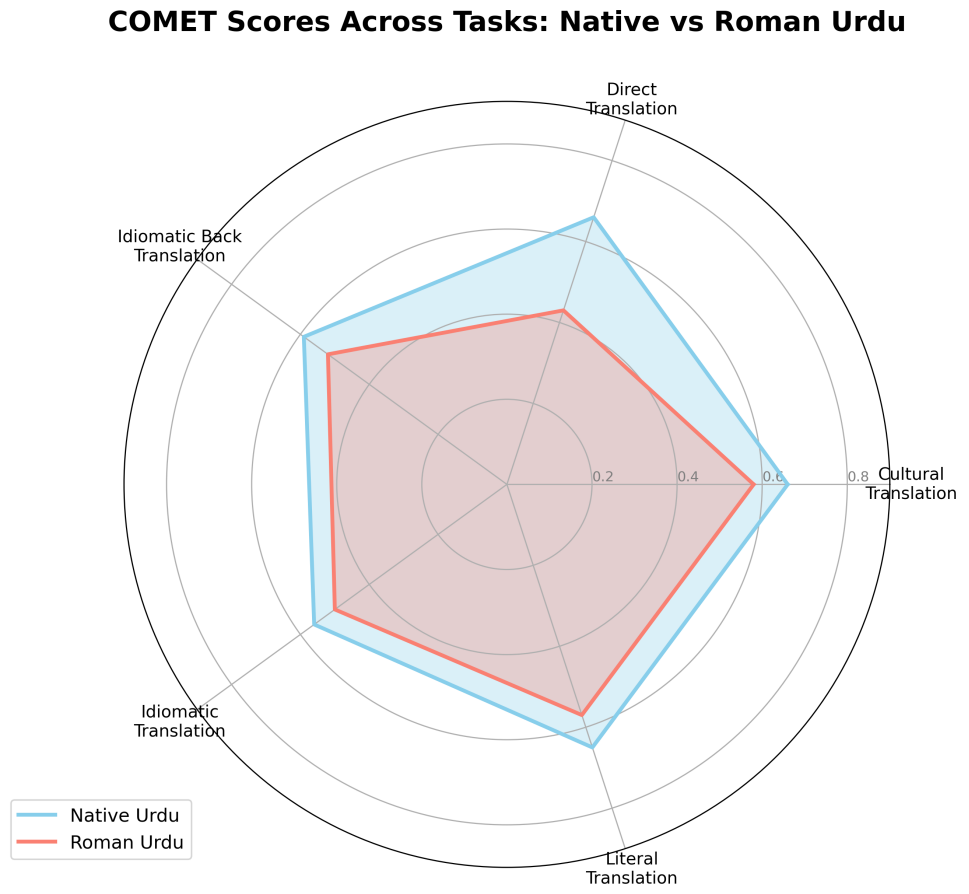

Figure 4: COMET Score Comparison: Performance across translation tasks for Native and Roman Urdu.

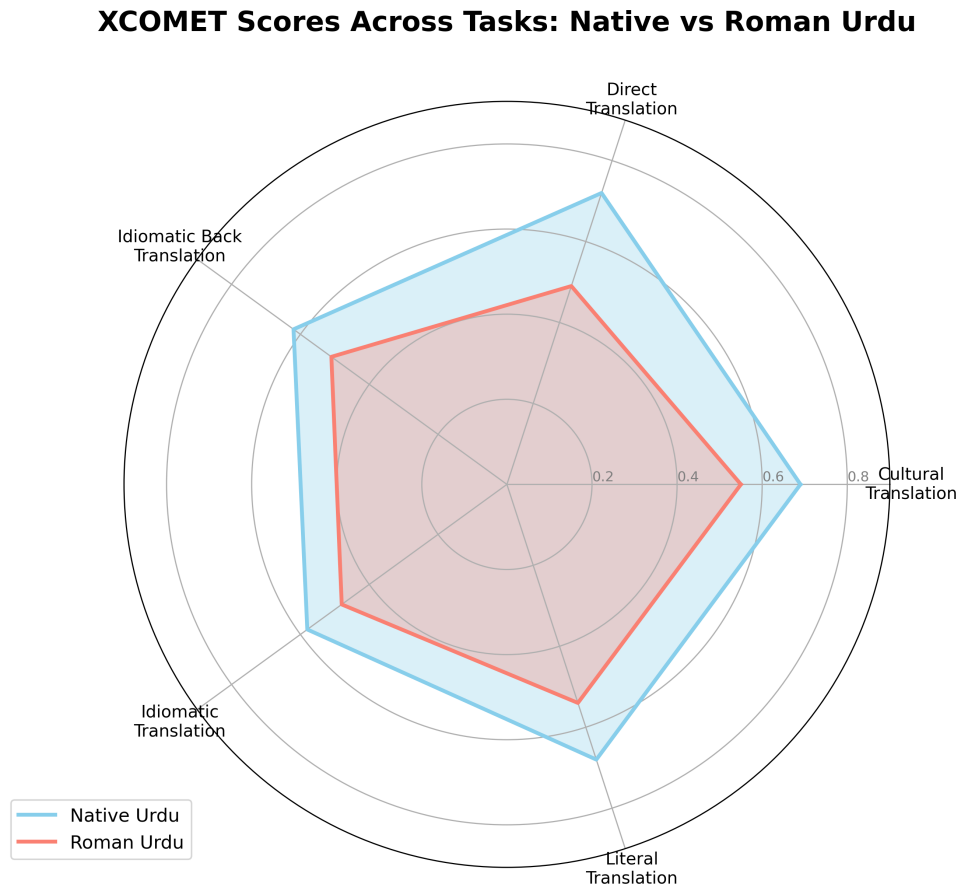

Figure 5: XCOMET Score Comparison: Performance across translation tasks for Native and Roman Urdu.